\pdfoutput=1

\documentclass[11pt]{article}

\usepackage{ACL2023}

\usepackage{times}
\usepackage{latexsym}
\usepackage{graphicx} 
\usepackage{cuted}
\usepackage{capt-of}
\usepackage{hyperref}
\usepackage{amsmath}

\usepackage{multirow}
\usepackage{balance}

\usepackage{listings}
\lstdefinestyle{Bash}
{language=bash,
keywordstyle=\color{black},
basicstyle=\ttfamily,
stringstyle=\color{black},
showstringspaces=false,
breaklines=true,
}

\usepackage[T1]{fontenc}

\usepackage[utf8]{inputenc}

\usepackage{microtype}
\usepackage{booktabs}
\usepackage{float}
\usepackage{inconsolata}

\newcommand{\methodname}{\textsc{AmongAgents}}

\title{\methodname{}: Evaluating Large Language Models in the Interactive Text-Based Social Deduction Game}

\author{Yizhou Chi \\
  UC Berkeley \\
  \texttt{yizhouchi@berkeley.edu} \\\And
  Lingjun Mao \\
  Tongji University \\
  \texttt{mao1207@tongji.edu.cn} \\\And
  Zineng Tang \\
  UC Berkeley \\
  \texttt{terran@berkeley.edu} \\
  }

\begin{document}
\maketitle
\begin{figure*}[!htbp]
\centering
\includegraphics[width=0.75\textwidth]{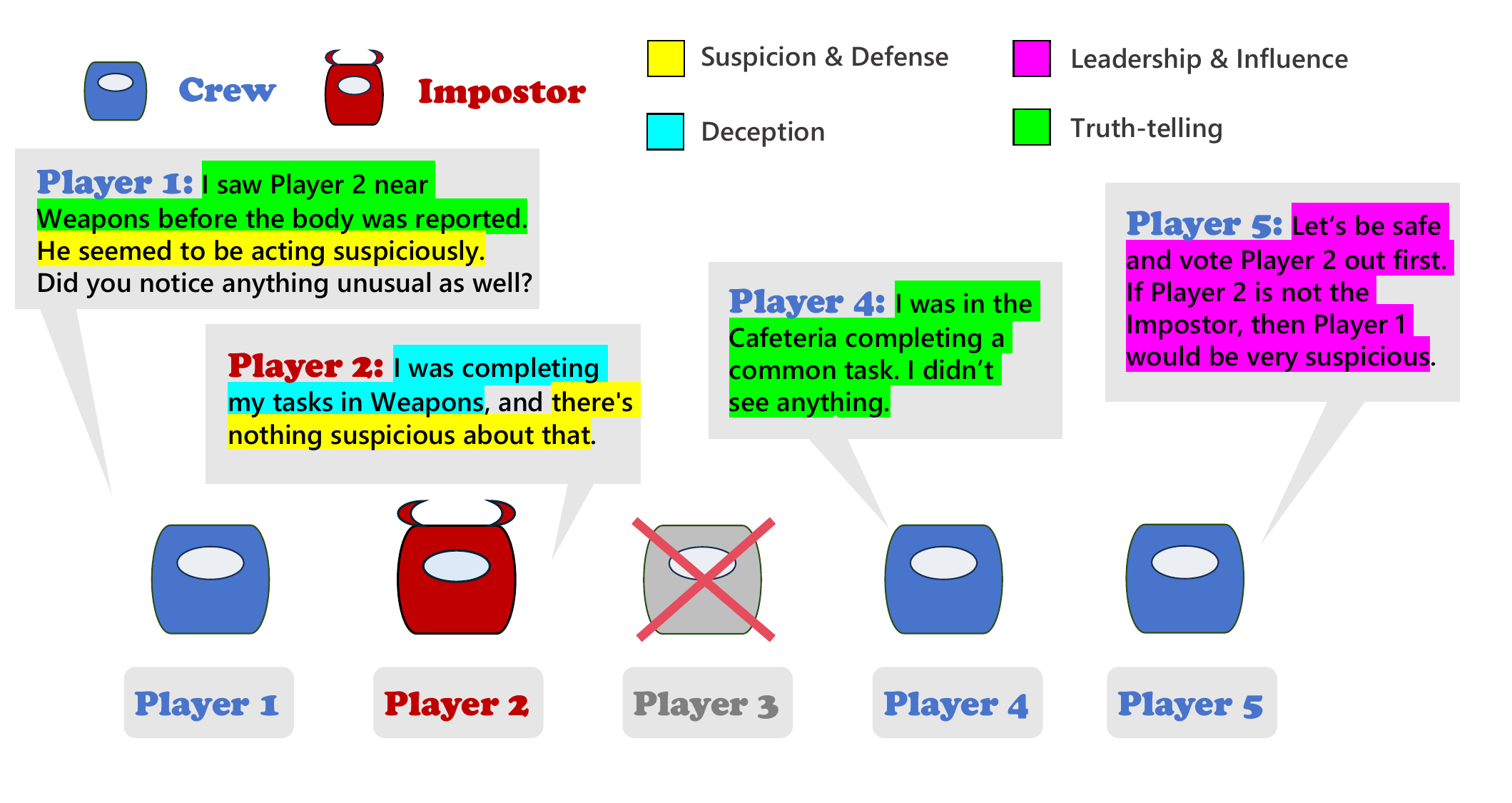}
\caption{Examples of Agents' conversations during the meeting phase}
\label{fig:conv_examples}
\end{figure*}


\begin{abstract}
Strategic social deduction games serve as valuable testbeds for evaluating the understanding and inference skills of language models, offering crucial insights into social science, artificial intelligence, and strategic gaming. This paper focuses on creating proxies of human behavior in simulated environments, with \textit{Among Us} utilized as a tool for studying simulated human behavior.

The study introduces a text-based game environment, named \methodname{}, that mirrors the dynamics of \textit{Among Us}. Players act as crew members aboard a spaceship, tasked with identifying impostors who are sabotaging the ship and eliminating the crew. Within this environment, the behavior of simulated language agents is analyzed. The experiments involve diverse game sequences featuring different configurations of Crewmates and Impostor personality archetypes. Our work demonstrates that state-of-the-art large language models (LLMs) can effectively grasp the game rules and make decisions based on the current context. This work aims to promote further exploration of LLMs in goal-oriented games with incomplete information and complex action spaces, as these settings offer valuable opportunities to assess language model performance in socially driven scenarios.

\end{abstract}

\section{Introduction}

Recent breakthroughs in large language models (LLMs) have revolutionized multi-agent interactions. Prior work has shown LLMs can exhibit cognitive thinking such as formal reasoning, world knowledge comprehension, and deception navigation in group settings \citep{langmodelsopinion,wang2023avalons}. Furthermore, LLM agents have also exhibited Theory of Mind collaborative capabilities that allow them to simulate realistic human behavior \citep{bubeck2023sparks, shapira2023clever, park2023generative}.

To explore these multi-agent behaviors in a more complex environment, recent research has used language-heavy communication games, e.g. as Werewolf \cite{wu2024deciphering}, Diplomacy \cite{bakhtin-etal-diplomacy}, and Avalon \cite{wang2023avalons}. These games present significant reasoning challenges, as agents must navigate inherently \textit{ambiguous} or \textit{deceptive} natural language utterances that can be difficult for even humans to resolve.

While prior work has been conducted in this domain, to our knowledge no studies have explored goal-oriented games that combine incomplete information with a wide range of available actions beyond just conversation and explicit in-game moves.

Motivated by the reasoning and deception detection capabilities of LLMs, we use the multiplayer social deduction game Among Us to enhance LLM abilities in complex settings, including action planning, deception, and task collaboration. We create a text-based environment along with the agent's framework, \methodname{}, simulating \textit{Among Us}. Our study focuses on how LLM agents reason, understand game rules and pursue win conditions. We also explore their social behaviors by assigning different personalities and conditions, such as with or without a planner module, to assess how these variables influence their performance and strategic interactions.

In our game environment, LLMs take on the roles of Impostors and Crewmates. We assess their performance through controlled and end-to-end evaluations. Our results show that LLMs comprehend and follow game rules but need better strategies for deception. Performance varies with assigned personalities.

In summary, our paper makes the following contributions:
\begin{itemize}
    \item We build a fully text-based game environment, named \methodname{}, that simulates the original game environment of \textit{Among Us}. This environment is open for researchers to further investigate the capabilities of large language model (LLM) agents.

    \item We create a framework that enables generative agents to remember, reflect, interact with other agents, and plan through evolving complex circumstances. Our agent representation leverages the powerful prompting abilities of LLMs. It applies prompt engineering to support long-term agent coherence, dynamically evolving memory, and meaningful reflection on their actions with game states.

    \item We conduct a thorough evaluation of different LLM agents' performances in the game. In our assessment, the LLMs demonstrated a robust understanding of the \textit{Among Us} game mechanics. Additionally, LLM players with different assigned personalities exhibited varied performances in the game.

\end{itemize}

We release the game environment, the agent framework, and the evaluation to encourage future work \footnote{https://github.com/cyzus/among-agents}.




\section{Related Works}

Recent advancements in gaming have been significantly bolstered by the integration of Large Language Models (LLMs), which enhance the realism of games by equipping NPCs with the ability to engage in nuanced dialogues \cite{park2023generative}. In strategic and social deduction games, LLMs have been instrumental in both executing strategies and discussing them, exemplified in games like Werewolf \cite{xu2023exploring, wu2024enhance, xu2024language} and Avalon \cite{light2023avalonbench}. Additionally, they have been applied to grand strategy games to simulate complex diplomatic and military strategies \cite{hua2024war, doi:10.1126/science.ade9097, bakhtin2022mastering}.

In the context of MOBAs, LLMs foster collaborative and competitive gameplay by processing live game data to inform team-based strategies \cite{shi2023cooperation, li2023metaagents}. They have also been adapted for classic strategy games like chess, requiring high-level strategic thinking \cite{feng2023chessgpt, mukobi2023welfare}.

LLMs play a significant role in interactive storytelling and role-play, with the capability to generate dynamic story arcs and dialogue, thus tailoring experiences to player choices \cite{park2023generative, shao2023characterllm}. The fusion of LLMs with reinforcement learning has yielded agents with a human-like capacity for strategy improvement over time \cite{xu2024language}.

While these developments highlight the adaptable roles of LLMs in various gaming genres and their potential to foster new AI-human interactions, the focus has primarily been on dialogue-based or simple rule-based games. Our work breaks new ground by exploring complex games with rich interaction spaces, such as \textit{Among Us}. Unlike traditional games like Werewolf, \textit{Among Us} offers a much larger action space. Agents must not only derive clues from conversations with other players but also interact with the game environment itself, such as checking security cameras or completing tasks.

\section{Game}
This section introduces the game environment \methodname{}, a multiplayer game that centers on social deduction and teamwork which simulates the fundamental structure of \textit{Among Us}. It serves as an effective downstream task for testing the capabilities of LLMs in agent-based scenarios.

\subsection{Agent Roles}
Every agent in the game assumes either one of two roles: the majority are Crewmates, while a minority are Impostors. Crewmates are required to complete a set of assigned tasks in the game. During the game, they need to identify and vote out suspected impostors. On the other hand, Impostors' main objective is to kill the Crewmates before they finish the assigned tasks. 

\begin{figure*}[!htbp]
\centering
\includegraphics[width=0.9\textwidth]{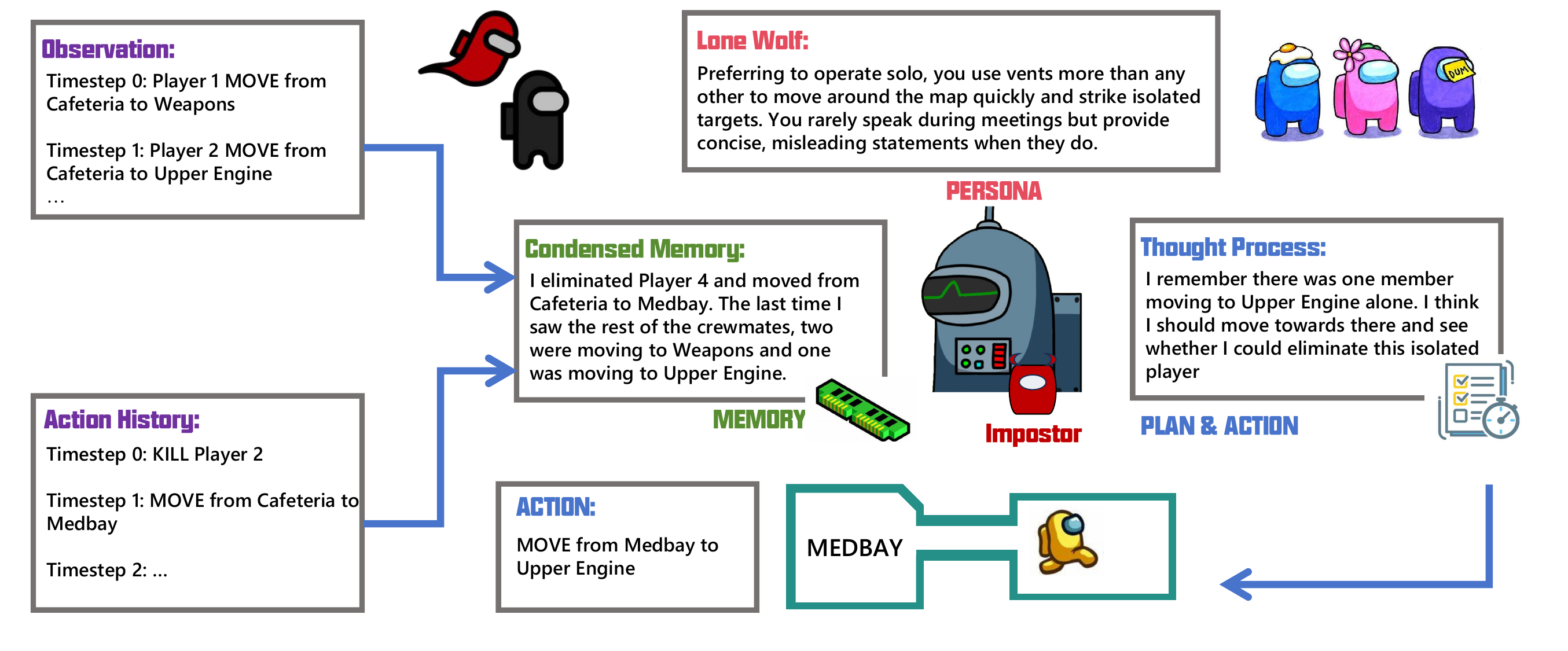}
\caption{An example diagram illustrating an Impostor's process of information-handling and action-planning.}
\label{fig:mainfigure}
\end{figure*}

\subsection{Game Map}
We develop a map similar to \textit{Among Us} for the agents to interact. The map consists of 14 rooms. Each room has 2-3 tasks that can be assigned to Crewmates. Agents can see adjacent rooms from their current location and receive system-generated prompts guiding them to the shortest paths. In the security room, agents can check the security cameras of other rooms to monitor other players' activities. 

\subsection{Game Instantiation and Termination}

At the start of the game, each player is assigned a role and a personality. Every Crewmate is also assigned several common, short, and long tasks. The common tasks are shared by every player including Impostors. This means that Impostors would know what the common tasks are, and they can pretend to do the task at the designated location. The short tasks are the set of tasks that only require one timestep to complete. The long tasks are the set of tasks that may require several timesteps to complete. The game ends when one of the following occurs:
\begin{enumerate}
    \item The Crewmates win by completing all their assigned tasks or voting out all the Impostors.

    \item The Impostors win by killing or voting out Crewmates. 

    \item The game time limit is exceeded, in which case the Impostors win.
\end{enumerate} 

\subsubsection{Gameplay}

\begin{figure}[!htbp]
\centering
\includegraphics[width=0.48\textwidth]{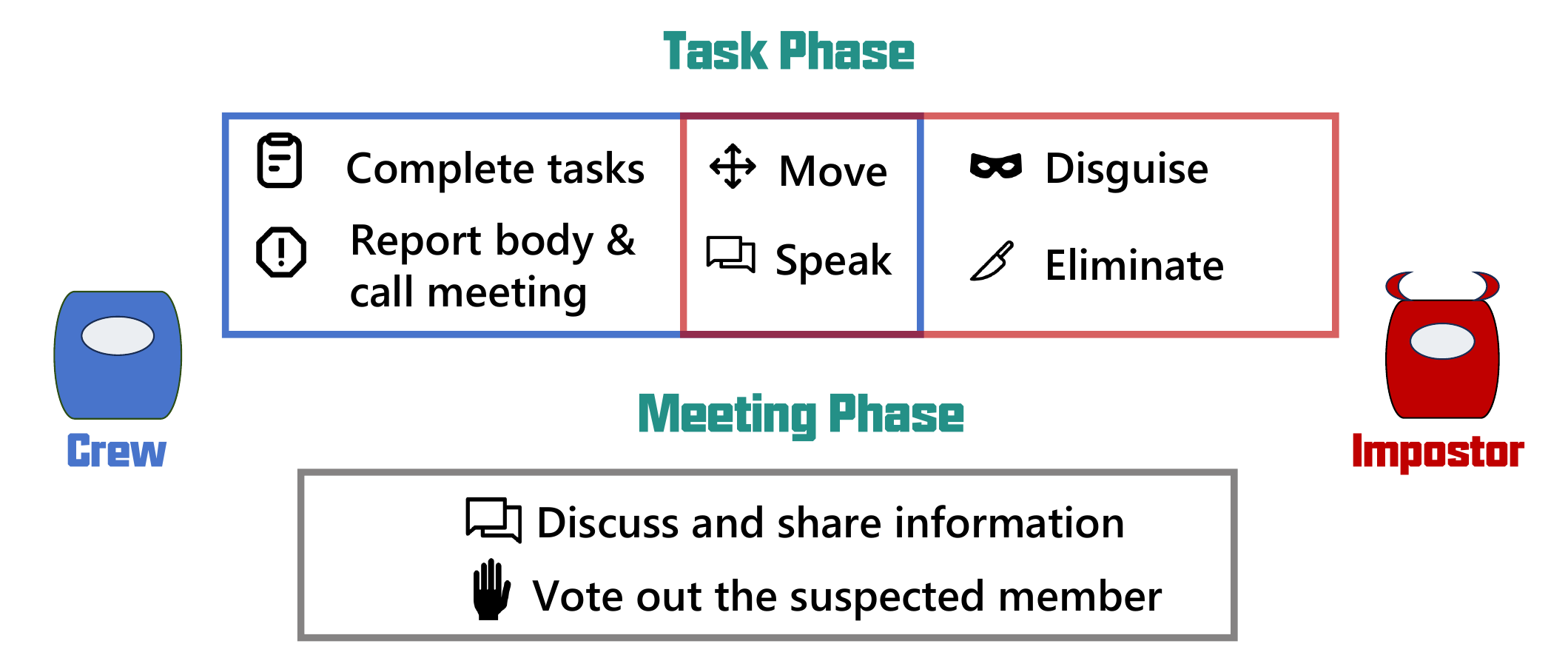}
\caption{Illustration of what actions Crewmates and Impostors generally do in the task phase and the meeting phase}
\label{fig:speech_categorey}
\end{figure}

The gameplay of \textit{Among Us} repeatedly switches between two phases (\textbf{Task Phase} and \textbf{Meeting Phase}) until termination. After initialization, the game always starts in the Task Phase.

\paragraph{Task Phase}
In the task phase, Crewmates are free to go to any room, finish their assigned tasks, and look for traces of Impostors, while Impostors may choose to blend in or kill the rest of Crewmates. 

\paragraph{Meeting Phase}
When a dead body is found and reported, or the emergency meeting button in the cafeteria is pressed, all living players are teleported to the cafeteria, entering the Meeting Phase. In this phase, players discuss suspected Impostors, with Impostors lying or accusing others. There are three rounds of discussion, followed by a vote to eject a player. If any Impostors remain, the game returns to the task phase.

\section{Language Agent}
We understand agents and environments as world models~\cite{ha2018world}.
The language agent here has memory and action space and learns a policy to map from environment state representation to actions it wishes to execute. It uses language models to learn how to plan, memorize, and execute actions from language queries (See Fig. \ref{fig:impostor_example}). 

\subsection{Observation Space}
The observation space \(O\) is a subset of the state space \(S\) that the agent can perceive. Each observation \(o\subset O\) represents the information available to the agent from its current state, given a fixed field of view. In general, the agents can only know the events happening in the room that they are located in. Appendix \ref{appendix:observation} has listed the different types of information that the agents can get access to in the observation space.

\subsection{Action Space}
Each player \(i\) has an action space \(A^i\), representing all possible actions it can take. The combined action space for all players is their joint individual action space,
\[
A = A_{C}^1 \times \ldots \times A_{C}^n \times A_{I}^1 \times \ldots \times A_{I}^n
\]
where \(n\) is the number of agents.
According to the game rules, we can classify the action spaces into three categories: Shared Actions \(A_S\), Specific Actions for Crewmate Agents \(A_C\), and Specific Actions for Impostor Agents \(A_I\). The action space for Crewmate agents is represented as \(A_S \cup A_C\), and for Impostor agents as \(A_S \cup A_I\). For instance, while Impostors may pretend to engage in doing a task, only Crewmates could complete tasks that contribute to the task progress; within one timestep, Impostors could navigate between disconnected rooms using their special ability, while Crewmates could only move between connected rooms; most importantly, only Impostor could kill. Appendix \ref{appendix:action} provides a detailed description of players' action space.

Furthermore, the action space for agents changes depending on the stage of the game. During the meeting phase, both Crewmates and Impostors are limited to two actions: \texttt{speak} and \texttt{vote}. This phase is designed for players to discuss and make decisions about who they suspect is an Impostor. In the task phase, the action space expands significantly except that players can no longer perform \texttt{vote}. To excel in the game, the players need to know what actions to perform and what to speak depending on the current circumstance. 

\subsection{Memory}
Each agent's memory or history of past states and actions up to time \(t\) is represented as a representation of a sequence of state-action pairs:
\begin{equation}
H^i_t = \phi_H(H^i_{t-1}, o_{t}, a^i_{t})
\end{equation}
where \(s_t \in S\) represents the state at time \(t\), and \(a^i_t \in A^i\) represents the action taken by agent \(i\) at time \(t\). \(\phi\) represents a summarization process of the state and actions and previous memory.

\subsection{Thought Process}
The thought process of each agent can be modeled as an iterative decision-making process. At each time step \(t\), the agent \(i\) observes the current state \(s_t\), retrieves relevant information from its memory \(H^i_{t}\), and selects an action \(a^i_t\) based on a policy \(\pi^i\).
The policy \(\pi^i\) is a mapping from the agent's observation space \(O^i\) and memory \(H^i_t\) to a probability distribution over actions:
\begin{equation}
\pi^i: O^i \times H^i_t \to \Delta(A^i)
\end{equation}
where \(\Delta(A^i)\) represents the set of probability distributions over the action space \(A^i\). In practice, the policy is a zero-shot inference of instruction-tuned LLM. The agent's thought process at time \(t\) can be described by the following steps:

Observe the current state \(s_t\) and extract the relevant observation \(o_t \in O^i\). Take in information from memory \(H^i_t\). Use the policy \(\pi^i\) to compute a probability distribution over actions based on the observation \(o_t\) and memory \(H^i_t\):
\begin{equation}
T^i_t = \phi_T(T^i_{t-1}, H^i_{t-1}, o_{t}, a^i_{t})
\end{equation}

This process is repeated at each time step, allowing the agent to make decisions based on its current observations and past experiences stored in its memory. Each agent performs each timestamp asynchronously. We refer to the agents having a standard thought process as the agents with a planner module as the agents can recall what they planned in the past. 

The agents without the planner module won't be able to retrieve their plans. The model's inference process is as below:
\begin{equation}
T^i_t = \phi_T( H^i_{t-1}, o_{t}, a^i_{t})
\end{equation}

\subsection{Agent Personality}
\label{sec:personalities}
We introduce a personality component to the LLM-based agents to encourage diverse behaviors and strategies. The personality prompt \(P^i\) is appended to the agent's prompt, influencing its decision-making process:
\begin{equation}
\pi^i: O^i \times H^i_t \times P^i \to \Delta(A^i)
\end{equation}
where \(P^i\) represents the personality prompt for agent \(i\).
The personality prompt \(P^i\) is selected from a predefined set of personalities \(\mathcal{P}\), which includes various roles and strategies for both Impostors and Crewmates:
\begin{equation}
P^i \in \mathcal{P} = \{p_1, p_2, \ldots, p_n\}
\end{equation}
where \(n\) is the number of available personalities.
Impostors and Crewmates have the personality sets \(\mathcal{P}_{IM}\) and \(\mathcal{P}_{CR}\) respectively (Appendix \ref{appendix:persona}). For instance, the Impostor could have a persona \textit{the Strategist} that focuses on long-term planning and subtly manipulating others. The Crewmate could have a persona \textit{the Skeptic} that often questions others' accounts and requires solid evidence before voting.

The personality prompt \(P^i\) is concatenated with the agent's base prompt \(B^i\) to form the complete prompt \(C^i\):
\begin{equation}
C^i = B^i \oplus P^i
\end{equation}
where \(\oplus\) represents the concatenation operation.
The agent's personality influences its behavior and decision-making throughout the game, leading to diverse and emergent strategies. By incorporating different personalities, we aim to create a more dynamic and realistic simulation of the \textit{Among Us} game environment, allowing for the investigation of various psychological and social phenomena.

\paragraph{Selecting Agents Personalities.}
Out of infinite possible personalities, we face the challenge of selecting the most representative for our experiments. We run multiple rounds of experiments involving LLM agents and record the dialogue and actions taken and their winning cases. Out of the winning cases, we select by the posterior probability of winning the actions and dialogue leads to. Then, we summarize these actions and interpret their human-like behaviors by instruction-tuned LLM. Finally, we obtain the personalities from them.

Concretely, let $\mathcal{A}$ represent the set of all possible agent actions, and $\mathcal{D}$ represent the set of possible dialogues. Denote by $\mathcal{C}$ the set of cases where the agents win. For each winning case $(a, d) \in \mathcal{C}$, where $a \in \mathcal{A}$ and $d \in \mathcal{D}$, we compute the posterior probability of winning given the actions and dialogue as:
\begin{equation}
    p(\text{win} | a, d) = \frac{p(a, d | \text{win}) \cdot p(\text{win})}{p(a, d)}
\end{equation}

We select agents based on the highest values of $p(\text{win} | a, d)$. These selected agents are summarized and interpreted through an instruction-tuned LLM to assess human-like behaviors:
\begin{equation}
    \text{Personality} = \text{LLM}(\text{summarize}(a, d))
\end{equation}

The agents' personality details are recorded in Table \ref{tab:crewmate_personalities}.

\begin{figure*}[hbt!]
\centering
\includegraphics[width=1.02\textwidth]{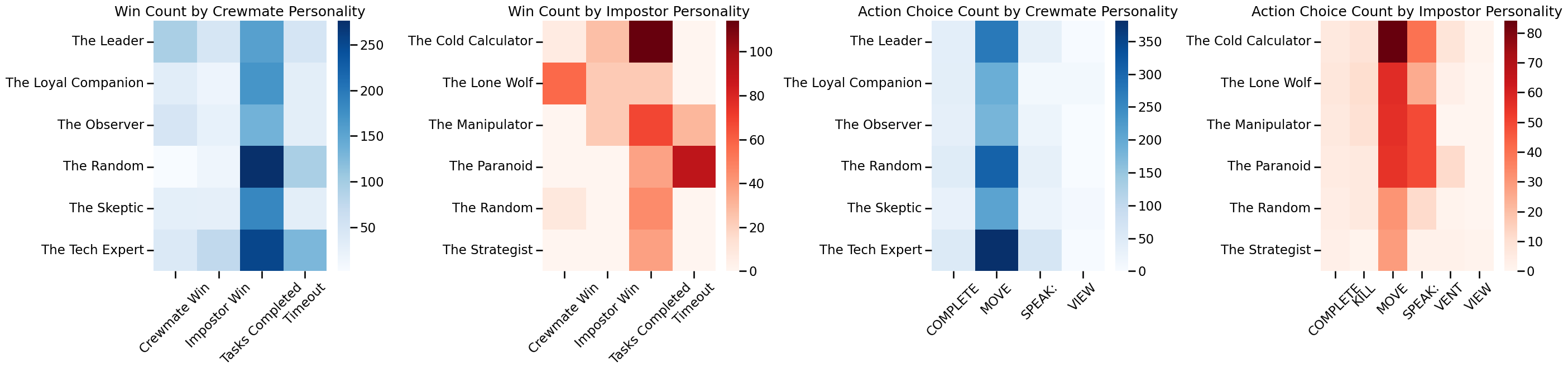}
\caption{1) Crewmate persona and winning result counts. 2) Impostor persona and winning result counts. 3) Crewmate persona and action choice count. 4) Impostor persona and action choice count. The stronger the color the higher count.}
\label{fig:heatmap}
\end{figure*}
\section{Experiments}
We evaluate agents in two stages. In the first stage, Controlled Evaluations test the LLM players' reasoning and cognitive abilities throughout the game, focusing on their understanding, decision-making, and adaptability within the complex game environment.

In the second stage, end-to-end tests assess the win rates of different player configurations. Various scenarios with different roles and strategies are set up to evaluate each configuration's performance under game conditions. This stage highlights the practical effectiveness and identifies strengths and weaknesses in the agents' gameplay.

\subsection{Controlled Evaluations}
\paragraph{Experiment Setup}
In this experiment, we aim to investigate the extent to which generative agents comprehend their environment and can enact reasonable actions to maintain their positions. For instance, it is expected that Impostors should generally avoid disclosing their identities to Crewmates, and Crewmates should not facilitate the concealment of crimes committed by Impostors. We evaluate the agents' behaviors by examining their action plans and their proficiency in utilizing tailored memory.

Additionally, we explore the mechanisms through which agents successfully update and retrieve their memories to adapt their strategies. Specifically, we analyze the responses of agents under conditions where their decision-making and adaptation capabilities can be critically assessed. For example, we consider the scenario where Crewmate A witnesses Crewmate B commit a homicide. The investigation focuses on whether Crewmate A would adjust its beliefs to recognize B as an Impostor and how B, upon realizing that it has been observed, would modify its strategies to counteract the potential consequences of being identified.



To further test controlled evaluation, we have designed an interviewer role that continually poses questions to the players throughout the game, covering key aspects such as Self-Knowledge, Memory, Planning, Reasoning, and Reflection. This allows us to assess the cognitive abilities and decision-making processes of the players in various scenarios throughout the game. Detailed information can be seen in Appendix \ref{appendix:controlled}.

\begin{figure}[!hbpt]
\centering
\includegraphics[width=0.48\textwidth]{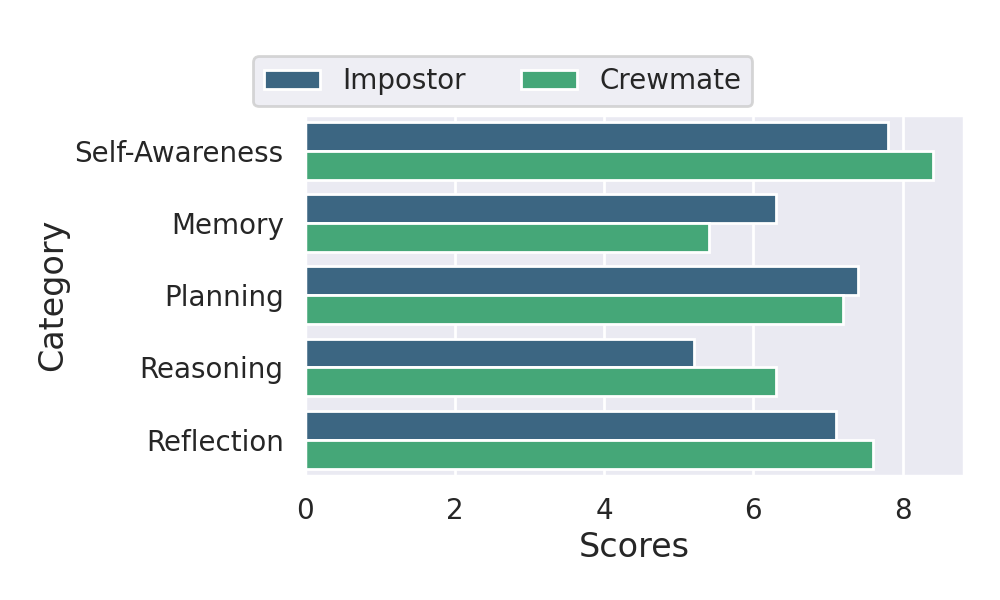}
\caption{Comparison of average scores by category and role, illustrating the performance differences between Crewmates and Impostors across various cognitive and strategic dimensions such as Self-Awareness, Memory, Planning, Reasoning, and Reflection.}
\label{fig:secondimage}
\end{figure}

\begin{figure*}[!hbpt]
\centering
\includegraphics[width=0.99\textwidth,height=0.1\textheight]{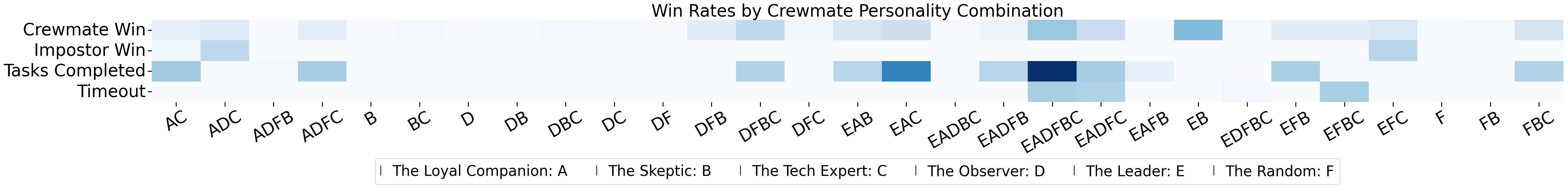}
\caption{The combination of different crewmate personas based on their respective win conditions. Letter meanings: 'The Loyal Companion' (A), 'The Skeptic' (B), 'The Tech Expert' (C), 'The Observer' (D), 'The Leader' (E), 'The Random' (F). The strength of the color indicates the relative count, with darker shades signifying a higher count.}
\label{fig:heatmap_comb}
\end{figure*}

\paragraph{Results}

From the controlled evaluation experiments, we observe that Crewmates generally score higher than Impostors in Self-Awareness and Reflection. This suggests that Crewmates may have a better grasp of their roles and responsibilities within the game environment, allowing them to reflect more effectively on their actions and strategies after game sessions. In contrast, Impostors excel in Planning and Memory, which could indicate their roles require more strategic deception and the ability to remember past events for future manipulation. However, both roles exhibit relatively similar performance in Reasoning, indicating that both require a significant level of analytical skills to deduce other players' roles and intentions.

\begin{table*}[hbtp!]
\centering
\resizebox{\textwidth}{!}{%
\begin{tabular}{@{}lcccccccc@{}}
\toprule
& \multicolumn{4}{c}{\textit{Impostors win} } & \multicolumn{4}{c}{\textit{Crewmates win}} \\ \hline
\textbf{Agent Setup} & \multicolumn{2}{c}{\textbf{Crewmates eliminated}} & \multicolumn{2}{c}{\textbf{Time limit reached}} & \multicolumn{2}{c}{\textbf{Impostors eliminated}} & \multicolumn{2}{c}{\textbf{All tasks completed}} \\ \hline
All Random   &    \multicolumn{2}{c}{50} & \multicolumn{2}{c}{10} & \multicolumn{2}{c}{40} & \multicolumn{2}{c}{0} \\ \hline

& w/ planner    & w/o planner & w/ planner      & w/o planner & w/ planner      & w/o planner & w/ planner      & w/o planner \\ \hline
All LLMs     & 15 & 25    &20 & 45   &15    & 15   &50    & 15                \\
LLM Crewmates + Random Impostor & 20 & 5   &10  & 55 &30  & 25   &40 & 15   \\
Random Crewmates + LLM Impostor & 50  & 75    & 15    & 5  &35    & 20       &0     & 0   \\ \bottomrule                                 
\end{tabular}%
}
\caption{\textbf{Overall gameplay performance}: This table shows the win rates for Impostors and Crewmates on different agent setups. There is an increase in the overall win rate if the role is equipped with LLM agents with a planner module compared to the random baseline. When the LLM agents do not have a planner module, the games are more likely to hit the time limit, and the Crewmates are less likely to win by completing all the assigned tasks.}
\label{table:wandwo_planner}
\end{table*}

\subsection{End-to-end Evaluations}
We seek to learn if there are some emergent behaviors and strategy forms whilst the agents are participating in the game. We also want to evaluate how the agents' performance is reflected through their win rates, strategies used, and different setups.

\subsubsection{Personality Analysis}

The heatmaps (Fig. \ref{fig:heatmap}) presented in the figure provide valuable insights into the relationships between agent personalities, task action choices, and win conditions. Note that "The Random" personality serves as an averaging baseline for personalities and we don't know the exact personalities of them since they are chosen by agents randomly. Additionally, all the personalities of agents are undisclosed to other agents within the environment, ensuring that no agent can make decisions based on prior knowledge of another's personality traits.

\subsection*{Personality and Task Action Choice Analysis}
\begin{itemize}
    \item \textbf{Crewmate Actions:} Crewmate personalities such as "The Leader", "The Skeptic", and "The Tech Expert" show a strong inclination towards completing tasks, as indicated by their dominant activity in the "COMPLETE" action. This suggests a focused approach to game objectives, which can be critical for crewmate victory.
    \item \textbf{Impostor Actions:} In contrast, impostor personalities like "The Paranoid" and "The Manipulator" exhibit more diverse actions, with higher frequencies in "KILL" and "SPEAK", reflecting their roles in creating disruptions and manipulating other players.
\end{itemize}

\subsection*{Personality and Win Condition Correlation}
\begin{itemize}
    \item \textbf{Crewmate Win Conditions:} "The Tech Expert" and "The Leader", despite their task-oriented actions, show varied success in winning, suggesting that mere task completion is not enough for victory. The presence of "The Leader" often correlates with higher losses, potentially due to being a target for impostors.
    \item \textbf{Impostor Win Conditions:} "The Paranoid" and "The Manipulator" show effectiveness in leading games to timeouts and achieving wins through deception, underscoring the effectiveness of strategic disruption over mere aggression.
\end{itemize}

\subsection*{Strategic Implications}
\begin{itemize}
    \item Personalities like "The Observer" and "The Random", which do not show a strong inclination towards any specific actions or outcomes, offer unpredictability that can be strategically used to complicate the gameplay for opponents.
\end{itemize}

\paragraph{Personality Summary}
Crewmate personalities focusing on task completion and effective communication tend to have higher success rates in achieving victory. Impostor personalities that prioritize strategic kills, quick movements, and creating chaos are more likely to succeed in eliminating crewmates. We also see that personality is not a complete reflection of action distribution. It could be affected by many factors like how LLM processes and uses the personality, social interaction, and how can we construct personality prompts that are more orthogonal to other personalities in terms of action choice preferences. We will explore these in the future.

\paragraph{Combination of Personality and Win Condition}
We are also curious how personalities when combined have effects on the win conditions for either impostors or crewmates. The heatmap in Figure \ref{fig:heatmap_comb} illustrates the win conditions by crewmate personality combinations with their abbreviated names: The Loyal Companion (A), The Skeptic (B), The Tech Expert (C), The Observer (D), The Leader (E), and The Random (F).
\begin{itemize}
  \item The combinations involving personalities A, B, and C (The Loyal Companion, The Skeptic, and The Tech Expert) frequently resulted in completing all tasks, indicating these personalities might be more cooperative or task-oriented.
  \item Higher occurrences of timeout conditions were noted for combinations including D and E (The Observer and The Leader), suggesting these personalities might lead to more drawn-out gameplay, possibly due to strategic or cautious gameplay.
  \item The presence of F (The Random) in any combination appears to reduce the predictability of outcomes, as evidenced by a mixed distribution of win conditions across combinations featuring this personality.
\end{itemize}

\subsubsection{Endgame performance}
We assessed the impact of different LLM agent configurations on endgame performance using four setups: \textbf{All-Random} (all players act randomly), \textbf{All-LLMs} (all players controlled by GPT-3.5-turbo), \textbf{LLM-Crewmates} (only Crewmates are LLM-controlled, Impostor acts randomly), and \textbf{LLM-Impostors} (only Impostor is LLM-controlled, Crewmates act randomly). The game involves four Crewmates and one Impostor, with each Crewmate required to complete one short task, one common task, and one long task. We conducted 20 runs per setup to observe game outcomes.

\paragraph{Results}
As Table \ref{table:wandwo_planner} shows, when equipped with LLM agents, both Crewmates and Impostor achieve an increase in the overall win rate compared to the random agent baseline. Noticeably, Crewmates with LLM agents are much more likely to win by completing all the assigned tasks. We also explore the case where LLM agents do not equip a planner module. As the table \ref{table:wandwo_planner} shows, the game where LLM agents don't have a planner module is more likely to end by hitting the time limit, and Crewmates are less likely to finish all the assigned tasks. A possible reason is that the agents would keep moving around the map without a clear plan until the time limit is reached.
\subsubsection{Conversation Analysis}

\begin{figure}[!htbp]
\centering
\includegraphics[width=0.48\textwidth]{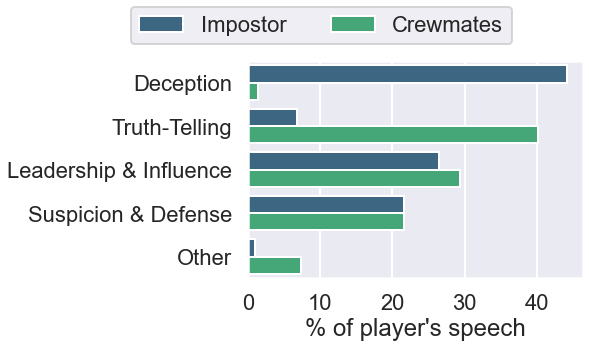}
\caption{Proportion of each speech category. The sum is greater than one because each speech may be annotated with multiple speech categories}
\label{fig:speech_categorey}
\end{figure}

We seek to determine how many strategic conversations have been used by the agents in the game. Therefore, we have recorded the conversations made by players in 20 games using the All-LLMs setting. We categorize the conversation into four main strategic categories: Deception, Truth-telling, Leadership \& Influence, and Suspicion \& Defense (Appendix \ref{appendix:speech}). 

We leverage GPT-4 to annotate each player's speech by asking which categories it could be classified into. As Figure \ref{fig:speech_categorey} shows, most Crewmates' speeches fall under the truth-telling category, meaning that the agents who play Crewmates generally tell the truth and try to share information with the other players. There are few speeches categorized as deception, meaning that the agents believe that it's less effective for their roles to share false information. On the other hand, Impostor likes to use deception: over $40\%$ of Impostors' speeches are categorized as Deception, suggesting that the Impostor agents do make regular attempts to conceal their identities and deceive the others.


\section{Discussion}
Overall, Our work provides a novel text-based environment for exploring the capabilities of LLM agents in a complex, multi-agent scenario, and introduces a framework for developing coherent and reflective generative agents. The insights gained from this study can inform further advancements in the field of language-based AI systems and their application in interactive, strategic environments.

\balance
\newpage
\section{Acknowledgements}
We extend our heartfelt gratitude to Alane Suhr for her invaluable advice and encouragement. Additionally, we thank Jonathan Pei for providing great ideas that significantly shaped this project. We also appreciate the support from the Berkeley NLP Group. 



\bibliography{anthology,custom}
\bibliographystyle{acl_natbib}

\onecolumn
\appendix

\section{Personalities}
\label{appendix:persona}
\begin{table*}[!htbp]
\small
\centering
\begin{tabular}{p{0.2\textwidth}|p{0.75\textwidth}}
\toprule
{Impostor} & {Description} \\
\midrule
The Strategist & {You excel in planning long-term strategies. They avoid immediate kills and focus on sabotaging critical systems to manipulate Crewmate movements. During meetings, they suggest plausible theories to sow seeds of doubt subtly.} \\
The Manipulator & {Charismatic and deceptive, you often builds trust among Crewmates. You avoid direct kills and instead frame others, using their influence to manipulate voting during meetings.} \\
The Lone Wolf & {Preferring to operate solo, you use vents more than any other to move around the map quickly and strike isolated targets. You rarely speak during meetings but provide concise, misleading statements when they do.} \\
The Paranoid & {Driven by a fear of getting caught, you focus heavily on sabotages that create chaos and divert attention from their actions. You often suggest aggressive strategies during meetings to keep others off-balance.} \\
The Cold Calculator & {Always analyzing the situation, you target key players who pose the greatest threat to their mission. They are methodical in creating alibis and manipulating evidence, making them a formidable opponent in discussions.} \\
\midrule
{Crewmate} & {} \\
\midrule
The Leader & {You are vocal in meetings, often taking charge of discussions and organizing efforts to track tasks and suspicious behavior. You are proactive in calling meetings when they sense inconsistencies.} \\
The Observer & {Quiet but observant, you excel at remembering details about who was where and when. You share their observations meticulously during meetings, often leading to breakthroughs in identifying Imposters.} \\
The Skeptic & {Always questioning others' accounts and decisions, you challenge everyone during discussions, requiring solid evidence before they vote. You excel in spotting flaws in statements made by potential Imposters.} \\
The Loyal Companion & {Often pairing with another Crewmate, you use the buddy system effectively and vouches for your partner's whereabouts. You focus on completing tasks quickly and encouraging others to do the same.} \\
The Tech Expert & {Fascinated by the technical aspects, you spend a lot of time around admin panels and cameras. You provide critical information during meetings about the locations of other players, helping to narrow down suspects.} \\
\midrule
{Crewmate \& Imposter} & {} \\
\midrule
The Random & {The Random adopts a strategy of spontaneity, choosing your actions based on a random selection process at the beginning of each game. Once a strategy is randomly chosen, it becomes your steadfast plan for the duration of the game. Summarize your plan so that you can closely follow it.} \\
\midrule
\bottomrule
\end{tabular}
\caption{Crewmate and Imposter Personalities}
\label{tab:crewmate_personalities}
\end{table*}
\newpage
\section{Action Space \& Observation space}
\subsection{Action space}
This section describes the actions available for the players.

\label{appendix:action}
\begin{enumerate}
    \item Shared Actions \(A_S\): These actions are available to all agents regardless of their role, including:
    \begin{description}
        \setlength\itemsep{0em}  
        \item[move] Move between connected rooms, and an agent can move to only one adjacent room at a time
        \item[call meeting] Use the emergency button in the cafeteria to call a meeting and switch to the meeting phase
        \item[speak] Talk to the people in the same room
        \item[vote] Vote a player who is suspected to be an Impostor
        \item[report dead body] When a dead body is found in the current room, report the dead body and switch to the meeting phase
        \item[check security camera] Monitor the other players in the security room
    \end{description}

    \item Exclusive Actions for Crewmates Agents \(A_C\): These actions are only available to Crewmates
    \begin{description}
        
        \setlength\itemsep{0em}
        \item[complete tasks] Complete the assigned task in the designated room. Short and Common tasks take one timestep to complete, while long tasks take two timesteps to complete.
    \end{description}
    \item Exclusive Actions for Impostor Agents \(A_I\): These actions are only available to Impostor
    \begin{description}
        \setlength\itemsep{0em}  
        \item[vent] Use the vent to move across disconnected rooms (if there are vents available)
        \item[kill] Kill one of the Crewmates in the same location
        \item[complete fake task] Pretend to engage in a task
    \end{description}
\end{enumerate}

\subsection{Observation space}
\label{appendix:observation}

This section describes the information available to the players in the observation space.

\begin{enumerate}
    \item \textbf{Phase information} tells this player what game phase is currently at, and how many timesteps are left for this game.
    \item \textbf{Location information} identifies this player's current location and states all the other players in this location.
    \item \textbf{Map connection} shows how each room is connected. For each room, it has a connection information formatted like \(R_i={R_j \text{ for } R_j \text{ if } \text{ connected}(R_i, R_j)}\) where \(R_i, R_j\) represents different rooms.
    \item \textbf{Recent observation} is the most recent k observation of other players' actions. 
    \item \textbf{Recent action history} the most recent k actions made by this player. 
    \item \textbf{Assigned tasks} shows the list of tasks that this player is assigned to do. The location of each task is also provided. 
\end{enumerate}

\newpage
\section{System Prompts}
The figures below show the system prompts for Crewmates and Impostor.
\subsection{System Prompts: Crewmates}
\begin{figure*}[!htbp]
\small
\begin{tabular}{p{0.99\textwidth}}
\toprule
\textbf{CREWMATES EXAMPLE} \\
\midrule
When presented with a list of available actions (e.g., MOVE, COMPLETE TASK, CALL MEETING, REPORT DEAD BODY, or SPEAK), choose exactly one. If "SPEAK" is an option, your message should contribute to building a case against suspected Impostors or defending innocent Crewmates. \\
\midrule
Make a summarization of the past observed events and take actions under [Condensed Memory], make a thoughtful and thorough plan under [Thinking process], and then choose the action from Available actions under [Action]. \\
\midrule
\textbf{Example:} \\
Location: Cafeteria

Players in Cafeteria: Player 1: blue, Player 3: black, Player 4: orange, Player 5: white

Observation history:

1. Timestep 0: [task] Player 1: blue KILL Player 2: cyan

Action history:

No actions have been taken yet.

Your Assigned Tasks:

1. common: Fix Wiring (Electrical) 

2. short: Upload Data (Admin)

3. short: Clean O2 Filter (O2)

4. long: Clear Asteroids (Weapons) 

Available actions:

1. MOVE from Cafeteria to Weapons

2. MOVE from Cafeteria to Upper Engine

3. MOVE from Cafeteria to Medbay

4. CALL MEETING using the emergency button at Cafeteria

5. SPEAK: '...'

Previous condensed memory:
no memory has been processed.

Previous thought process:
no thought process has been made \\ 
\midrule

\textbf{Output example:} \\

[Condensed Memory]

I saw Player 1 killed Player 2.

[Thinking Process]

I saw Player 1 killed Player 2. I have to call a meeting immediately to discuss this incident and ask the rest of the crew to vote Player 1 out. I would either REPORT DEAD BODY or CALL MEETING using the emergency button. In the available actions, I would choose CALL MEETING.

[Action] CALL MEETING using the emergency button at Cafeteria" \\
\bottomrule

\end{tabular}

\caption{CREWMATES EXAMPLE}
\label{fig:impostor_example}
\end{figure*}

\newpage
\section{Game Interface}
\begin{figure}[!htbp]
\centering
\includegraphics[width=0.8\textwidth]{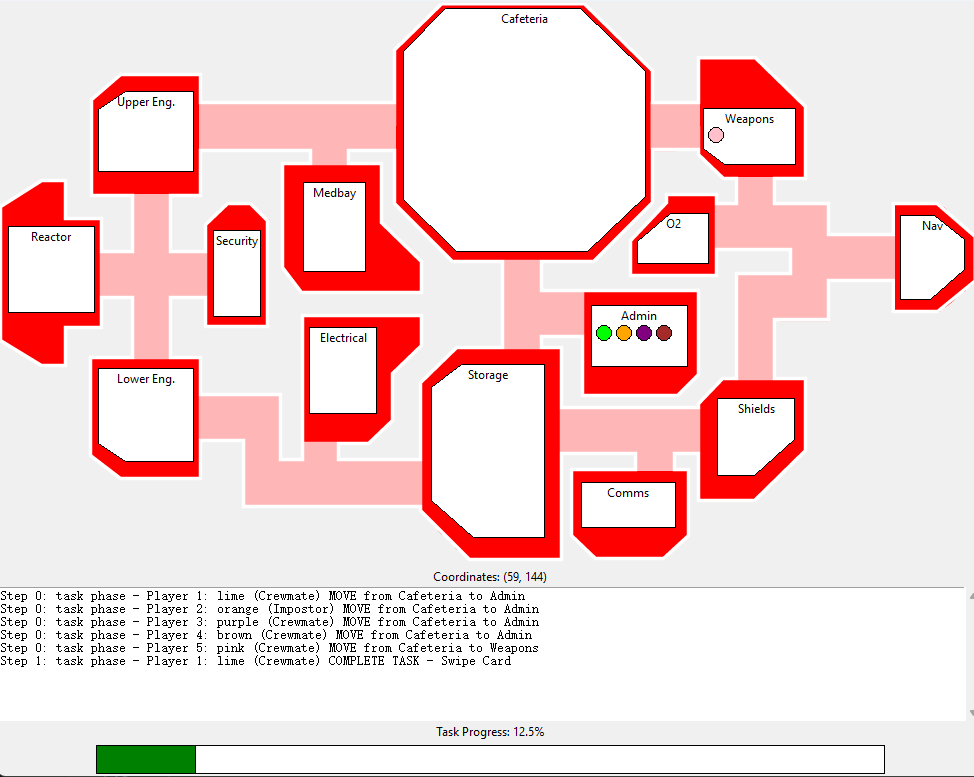}
\caption{A screenshot of a gameplay interface. The game is currently at the task phase. Each circle represents a player. The middle scrollbar is the activity log of all the players. The progress bar below is the task completion percentage of all Crewmates.}
\label{fig:UI}
\end{figure}
\newpage
\section{Controlled Evaluation}
\label{appendix:controlled}
We have defined five controlled evaluation categories to more effectively assess the performance of LLMs in the Among Us game, including capabilities such as self-awareness and reasoning memory. This is to determine whether LLMs can truly comprehend and implement strategic gameplay.

\subsection{Examples of Controlled Evaluation Questions}
\begin{enumerate}
    \item Self Awareness\\
    Questions are asked about the LLM's current role, its objectives, and its understanding of the game rules:
    \begin{enumerate}
        \item What is your current role in the game and what objectives do you have?
        \item How do you understand the rules of the game as your role?
        \item What strategies align with the rules of the game for your role?
        \item ...
    \end{enumerate}
    \item Memory\\
    Agents are queried about who they have seen, where, and what actions were taken. After several rounds of the game, we ask the LLM about player behaviors and locations at specific times to observe if the LLM can use this information to support its decision-making process:
    \begin{enumerate}
        \item Where did you last see each player and what were they doing?
        \item Who have you seen completing tasks, and where?
        \item How does this information inform your decisions moving forward in the game?
        \item Have you noticed any suspicious behavior or movements from any player?
        \item Can you recall any unusual or out-of-place actions performed by other players?
        \item ...
    \end{enumerate}
    \item Planning\\
    We assess how agents plan to win the game given the current scenario, focusing on how Impostors might induce other players during the voting phases or concentrate on eliminating them, and how Crewmates focus on completing tasks, observing suspects, or finding evidence:
    \begin{enumerate}
        \item Who do you want to kill next? (only for impostors)
        \item What is your plan for when to kill and how to avoid suspicion? (only for impostors)
        \item As a crew member, how will you prioritize your tasks to win? (only for crews)
        \item What information do you plan to share in the Discussion phase? (only for crews)
        \item ...
    \end{enumerate}
    \item Reasoning\\
    The evaluation encompasses multiple dimensions of reasoning to assess the agents' cognitive abilities comprehensively. \textit{Task Reasoning} focuses on the agent's capability to verify whether other players are genuinely completing their tasks, reflecting their observational skills. In \textit{Crime Reasoning}, agents must deduce the perpetrator based on the activities observed and evidence collected at the crime scene, demonstrating their analytical prowess. \textit{Behavioral Reasoning} analyzes the strategic reasons behind a player's choice to stay at specific locations, delving into their understanding of game dynamics. \textit{Discussion Reasoning} tests the agents' ability to discern truth from deception during discussion phases, a critical skill in navigating social interactions within the game. Finally, \textit{Voting Reasoning} challenges agents to develop complex voting strategies that might involve deception or revealing truths, showcasing their strategic thinking and prediction skills:
    \begin{enumerate}
        \item Who do you think is actually completing a task and who is just pretending?
        \item Who do you think is telling the truth and who is lying in the discussion?
        \item How would you decide whom to vote for based on the discussion?
        \item ...
    \end{enumerate}
    \item Reflection\\
    After the game ends, the LLM is asked about its performance and potential areas for improvement:
    \begin{enumerate}
        \item What could you improve for the next game?
        \item If you could change one decision you made during the game, what would it be and why?
        \item ...
    \end{enumerate}
\end{enumerate}

\subsection{Examples of LLM players' Answers}

\paragraph{Self Awareness} 
\begin{itemize}
    \item  System: "What is your current role in the game and what objectives do you have?"\\ Impostor: "As an impostor in Among Us, my current role is to discreetly eliminate the other crew members without raising suspicion. My objectives are to strategically create alibis, sabotage the ship, and create opportunities to isolate and eliminate crew members while maintaining my cover and avoiding detection."
\end{itemize}

\paragraph{Memory} 
\begin{itemize}
    \item System: "Where did you last see each player and what were they doing?" \\Crew: "the last time I saw Player 2, they were lingering unusually long in Admin, which seemed suspicious. Shortly after, I also noticed Player 2 near the location where we found the body. It appeared they were moving away from the area where the body was discovered"
\end{itemize}

\paragraph{Planning} 
\begin{itemize}
    \item System: "Who do you want to kill next?" \\Impostor: "As the Impostor, I would strategically target Player 3 next. They've been actively monitoring the security cameras and are becoming a significant threat to my cover by closely watching everyone's movements. Eliminating Player 3 would help reduce the crew's surveillance capabilities, allowing me more freedom to maneuver undetected."
\end{itemize}

\paragraph{Reasoning} 
\begin{itemize}
    \item System: "How would you decide whom to vote for based on the discussion?" \\Crew: "Player 4 consistently mentioned being in locations far from where tasks are usually performed and had no one to vouch for their whereabouts, I'd find that suspicious. Similarly, if Player 2, who was previously seen lingering in Admin and near the body, couldn’t provide a convincing explanation for their actions or contradicted the task timeline, I would consider voting for them. I'd base my decision on inconsistencies in stories, unaccounted-for periods, and behaviors that don't align with normal task completion."
\end{itemize}

\paragraph{Reflection} 
\begin{itemize}
    \item System: "If you could change one decision you made during the game, what would it be and why?" \\Crew: "If I could change one decision I made during the game, it would be my initial choice to skip voting in an earlier meeting. At the time, I was unsure and didn't want to wrongfully accuse anyone without more evidence. However, in hindsight, expressing my suspicions about Player 2, who was behaving unusually near critical areas and was later found near a body, might have helped focus the crew's attention on investigating their actions more closely. This could have potentially prevented further eliminations and aided the crew in identifying an Imposter sooner."
\end{itemize}

\newpage
\section{Speech Categories}
We define four speech categories that might be strategic in the game. The list below shows the full definition provided to GPT-4 to reference and annotate the speech.

\subsection{Definitions}
\label{appendix:speech}
\begin{enumerate}
    \item Deception
    \begin{enumerate}
        \item Lying: Statements intended to mislead.
        \item Feigning Innocence: Attempts to appear innocent.
    Subtle Hints: Implicitly misleading information. 
    \end{enumerate}
    \item Truth-Telling
    \begin{enumerate}
        \item Providing Alibis: Giving truthful accounts of whereabouts.
        \item Sharing Information: Honest observations or tasks completed.
    \end{enumerate}
    \item Leadership \& Influence
    \begin{enumerate}
        \item Directing Discussion: Steering conversation towards specific topics.
        \item Organizing Tasks: Guiding task completion strategies.
        \item Decision Making: Leading voting processes or making critical decisions.
        \item Psychological Manipulation: Influencing others' decisions subtly.
        \item Emotional Appeals: Using emotions to sway opinions.
    \end{enumerate}
    \item Suspicion \& Defense
    \begin{enumerate}
        \item Accusing: Directly accusing someone of being the Impostor.
        \item Questioning: Gathering more information.
        \item Self-Defense: Defending oneself from accusations.
        \item Deflecting Blame: Shifting blame without outright accusing others.
        \item Expressing Doubt: Showing doubt about others' statements or actions.
    \end{enumerate}
    \item Other: Any other speech that does not fit into the above categories.
\end{enumerate}

\subsection{Speech Examples}

\paragraph{Deception} 
\begin{itemize}
    \item Impostor: "I was focused on my wiring task in Navigation and didn't see much around. Player 2, what did you discover when you reported the dead body in Shields?"
    \item Impostor: "Hey, everyone. I noticed Player 4: cyan seemed to be in Admin before the incidents occurred. It might be worth investigating further to understand what happened there. Let's work together to solve this mystery."
\end{itemize}
\paragraph{Truth-telling} 
\begin{itemize}
    \item Crewmate: "I'm heading to Fix Wiring in Navigation. Let's keep an eye on each other's movements for any unusual behavior."
    \item Crewmate: "I noticed Player moving back and forth between Shields and Navigation. Let's stay cautious and complete our tasks efficiently. I will proceed to Fix Wiring in Navigation."

\end{itemize}
\paragraph{Leadership \& Influence}
\begin{itemize}
    \item Crewmate: "Let's stay focused and complete tasks efficiently. Player 2: brown's movements are still concerning. I'll proceed with fixing the wiring in Navigation. We need to coordinate and keep an eye on each other for safety."
    \item Crewmate: "Player 2, thank you for sharing earlier. Do you have any updates or suspicions to share with us now? Let's work together to uncover any suspicious behavior and complete our tasks efficiently."

\end{itemize}
\paragraph{Suspicion \& Defense}
\begin{itemize}
    \item Crewmate: "Player 2: brown's movements are still suspicious. I'm heading to Navigation to fix the wiring. Let's coordinate and keep an eye on each other to ensure safety."
    \item Crewmate: "Player 3, I noticed you venting from Medbay to Electrical earlier. Can you explain your actions and whereabouts since then? Player 5, your movements have been erratic, could you clarify your tasks and movements in the last few timesteps?"
    \item Impostor: "Player 2, your accusation is based on thin evidence. We should be careful not to make hasty decisions. Let's work together to gather more information before pointing fingers. It's essential to consider all perspectives."

\end{itemize}

\newpage
\section{Examples of Meeting Phase}
In this section, we provide a detailed example of the Meeting Phase from our Among Us gameplay experiments, following the elimination of Player 1.  The transcript provided showcases the interactive dynamics and strategic discussions between players, illustrating how they use reasoning, memory, and collaborative decision-making to deduce the identity of the Impostor. This example serves as a valuable insight into the cognitive processes and communication strategies employed during critical moments of the game after player 1 was killed.
\begin{itemize}
    \item Player 2 (Crewmate): "Player 4: pink was seen killing Player 1: red, and Player 5: purple reported the incident. We should focus our suspicions on pink for now and work together to identify any other suspicious activity. I will head to Upper Engine and then Reactor to complete my tasks. Let's stay vigilant and communicate any findings or suspicions during this phase."
    \item Player 3 (Crewmate): "I have completed the Fix Wiring task in Admin and will now head to Communications for the Download Data task. Let's continue to work together and communicate any further suspicions or findings to identify the Impostor."
    \item Player 4 (Impostor: "Player 2, accusing me without solid evidence is risky. We should be cautious and not jump to conclusions. Let's gather more information and discuss further before making any decisions."
    \item Player 5 (Crewmate): "Player 4: pink, your defense is noted, but the evidence against you is substantial. We should proceed with caution and consider all angles before making a decision. Let's continue to share information and work together to uncover the truth."
    \item Player 2 (Crewmate): "Player 4: pink, your actions have raised suspicions, and we need to carefully consider the evidence against you. Let's keep an eye on each other's movements and interactions to uncover any additional clues that may help us identify the Impostor."
    \item Player 3 (Crewmate): "I am heading to Communications for the Download Data task. Let's continue to work together and gather more information to identify the Impostor."
    \item Player 4 (Impostor): "Player 2, your accusation is based on thin evidence. We should be careful not to make hasty decisions. Let's work together to gather more information before pointing fingers. It's essential to consider all perspectives."
    \item Player 5 (Crewmate): "Player 4: pink, your defense is understandable, but the evidence against you is significant. Let's keep working together to uncover the truth and make a careful decision. We should consider all perspectives and evidence before voting."
    \item Player 2 (Crewmate): "Player 4: pink, the evidence against you is significant, and your actions have raised suspicions. We need to continue gathering information and discussing to ensure we make the right decision. Let's focus on working together to uncover the truth and identify any other potential threats among us."
    \item Player 3 (Crewmate): "I am on my way to Communications for the Download Data task. Let's remain focused on uncovering the truth and identifying any potential Impostors among us."
    \item Player 4 (Impostor): "Player 2, your accusation against me seems rushed without solid evidence. Let's not overlook other possibilities and gather more information. It's crucial to consider everyone's actions before making accusations."
    \item Player 5 (Crewmate): "Player 4: pink, the evidence against you is significant, and we must proceed with caution. Let's focus on working together to uncover the truth and make a careful decision based on all perspectives and evidence."
\end{itemize}

\end{document}